\documentclass[11pt]{article}

\usepackage[final]{acl}

\usepackage{times}
\usepackage{latexsym}

\usepackage[T1]{fontenc}

\usepackage[utf8]{inputenc}

\usepackage{microtype}

\usepackage{inconsolata}
\usepackage{xcolor}
\usepackage{colortbl}
\definecolor{BlueGreen}{RGB}{13, 152, 186}
\definecolor{Green}{RGB}{0, 128, 0}
\definecolor{Gray}{RGB}{128, 128, 128}
\definecolor{Blue}{RGB}{0, 0, 255}
\definecolor{softblue}{RGB}{100, 149, 237}
\hypersetup{
  colorlinks=true,
  linkcolor=black,
  citecolor=lightblue,
  urlcolor=black
}
\usepackage{url}
\usepackage{booktabs}
\usepackage{lineno}
\usepackage{multirow}
\usepackage{graphicx}
\usepackage{amsmath}
\usepackage{mdframed}
\usepackage{lipsum}
\usepackage{multicol}
\usepackage{verbatim}
\usepackage[normalem]{ulem} 
\usepackage{wrapfig} 
\usepackage{adjustbox} 
\usepackage{placeins}

\usepackage{float}
\definecolor{darkblue}{rgb}{0, 0, 0.5}
\usepackage[table]{xcolor} 
\definecolor{HeaderBG}{HTML}{F2F4F8}
\definecolor{BandBG}{HTML}{FAFBFF}
\definecolor{OursBG}{HTML}{F7F9FC}
\definecolor{em}{gray}{0.9}
\definecolor{lightblue}{rgb}{0.22,0.45,0.70}
\newcommand{\cem}{\cellcolor{em}}
\title{INMS: Memory Sharing for Large Language Model based Agents}

\author{Hang Gao \\
  Rutgers University \\
  \texttt{h.gao@rutgers.edu} \\\And
  Yongfeng Zhang \\
  Rutgers University \\
  \texttt{yongfeng.zhang@rutgers.edu} \\}

\begin{document}
\maketitle
\begin{abstract}
While Large Language Model (LLM) based agents excel at complex tasks, their performance in open-ended scenarios is often constrained by isolated operation and reliance on static databases, missing the dynamic knowledge exchange of human dialogue. To bridge this gap, we propose the INteractive Memory Sharing (INMS) framework, an asynchronous interaction paradigm for multi-agent systems. By integrating real-time memory filtering, storage, and retrieval, INMS establishes a shared conversational memory pool. This enables continuous, dialogue-like memory sharing among agents, promoting collective self-enhancement and dynamically refining the retrieval mediator based on interaction history. Extensive experiments across three datasets demonstrate that INMS improves agent performance by effectively modeling multi-agent interaction and collective knowledge sharing.
\end{abstract}

\section{Introduction}
The emergence of Large Language Model (LLM) based agents has significantly advanced machine learning and conversational AI. In particular, the advent of In-Context Learning (ICL) \citep{brown2020language} has enabled LLM-based agents to perform tasks using natural language prompts and few-shot examples without explicit retraining or fine-tuning, thus broadening their applicability across diverse domains \citep{ahmed2022few, izacard2022atlas}. Building upon ICL, the development of Chain-of-Thought (CoT) prompting further enhanced agents' ability to handle complex reasoning and arithmetic tasks \citep{wei2022chain}. Recent advances such as PAL \citep{gao2023pal}, integration with symbolic solvers \citep{he2023solving}, and continuous skill acquisition \citep{wang2023voyager} have further improved LLM-based agents' reasoning and adaptability. Despite these advancements, generating desired responses for open-ended questions remains challenging when agents operate in strict isolation. Human intelligence highly relies on social interaction and continuous dialogue to construct collective knowledge; however, most current agents lack this interactive dynamic, restricting their performance to the limited scope of isolated reasoning and statically provided examples.

To address this challenge, Retrieval-Augmented Generation (RAG) \citep{lewis2020retrieval} has been introduced, significantly increasing the availability of relevant examples for open-domain queries \citep{mao2020generation, gao2026beyond}. Additionally, self-learning mechanisms have been integrated with retrieval methods to dynamically improve text generation quality \citep{rubin2021learning, wang2023learning}. However, RAG primarily functions as a one-way static information lookup rather than a dynamic conversational process. Its effectiveness heavily depends on the quality, availability, and up-to-date status of external databases. The scarcity and obsolescence of these static databases become even more pronounced in specialized and rapidly evolving fields, where knowledge is ideally constructed and refined through continuous, interactive discourse rather than rigid, isolated retrievals.
\begin{figure}[t]
\begin{center}
\includegraphics[width=1\columnwidth]{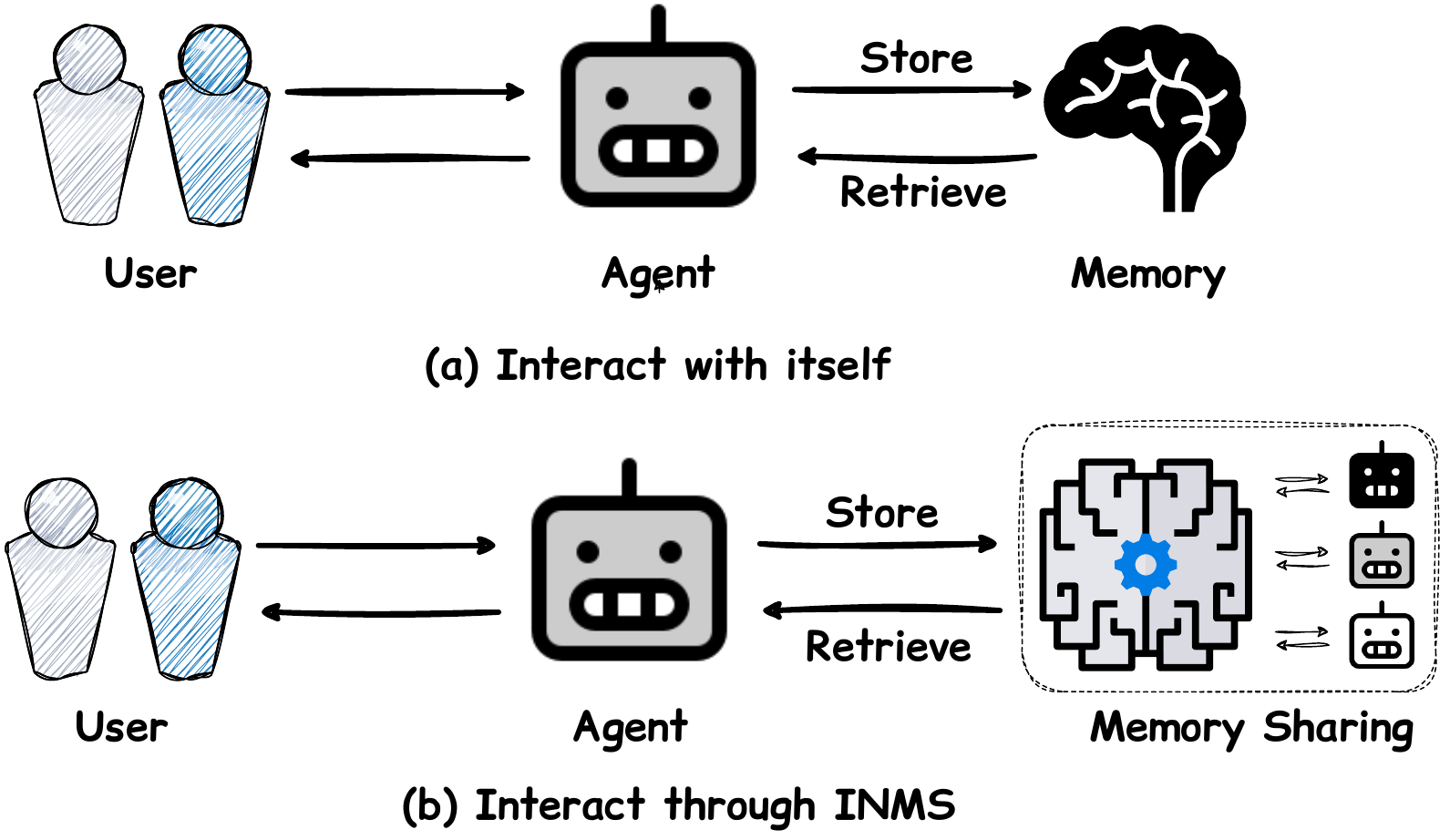}
\caption{(a) In traditional setups, each agent interacts only with its own memory, limiting knowledge reuse.
(b) In our proposed INMS framework, agents interact through a shared memory space that supports storing, retrieving, and cross-agent knowledge sharing, enabling collaborative improvement across agents.}
\vspace{-8mm}
\label{fig:figure3}
\end{center}
\end{figure}

Furthermore, existing memory enhancement methods primarily focus on agents independently utilizing stored historical information to inform responses. This isolated paradigm neglects the profound potential of inter-agent interactions and collective memory utilization. Current approaches often fail to model the asynchronous dialogue and knowledge exchange that naturally occurs in complex multi-agent environments, missing out on the inherent diversity and complementarity of agents that hold unique conversational histories and specialized training.

Motivated by the need to model effective multi-agent communication and overcome these limitations, we introduce the \textbf{IN}teractive \textbf{M}emory \textbf{S}haring (INMS) framework. INMS shifts the paradigm from isolated reasoning to an implicit, highly efficient asynchronous dialogue mechanism. By facilitating dynamic memory sharing through interactive learning, INMS establishes a shared conversational ground without reliance on external static databases. Figure.\ref{fig:figure3} illustrates the INMS workflow, emphasizing how this continuous, asynchronous interaction among agents enriches the shared memory pool and promotes the evolution from individual problem solvers toward an interactive, collectively intelligent society. Asynchronous interaction refers to memory-mediated exchange: agents contribute filtered Prompt-Answer pairs to a shared pool, and later agents reuse these memories as contextual examples. INMS is designed for open-ended, experience-accumulating scenarios in which agents possess different interaction histories and useful task-specific memories need to be constructed dynamically.

Specifically, within INMS, interactions between agents generate Prompt-Answer (PA) pairs, conceptualized as asynchronous communicative acts or interactive memories. These PA pairs are rigorously evaluated by a dedicated LLM scorer acting as a dialogue moderator, filtering out ambiguous or irrelevant content to maintain a high-quality interaction history. Accepted PA pairs continuously expand the shared memory pool. Crucially, an autonomous retriever dynamically updates and refines its capabilities based on this growing pool, acting as an adaptable mediator of this implicit dialogue.  This mechanism ensures that the most relevant experiential context is curated for each query, mitigating "echo chamber" biases from initial interactions and promoting progressively higher response accuracy and relevance.

Moreover, we constructed a novel dataset specifically targeting the multifaceted nature of open-ended tasks—such as poetry generation, unconventional logical problem-solving, and plan creation domains where interaction, nuance, and rich language generation are paramount. Through extensive experimentation across three diverse domains, each involving three specialized agents, we demonstrated that INMS enhances response precision and effectively models multi-agent communication. In summary, the primary contributions of this paper include:
\begin{itemize}
\item Introducing the INMS framework, an asynchronous interaction paradigm that leverages dynamically generated and shared conversational memories among agents, evaluated through rigorous scoring criteria, to enhance collective performance and retriever accuracy.
\item Addressing the limitations of static external datasets by modeling continuous, interactive memory generation, thereby facilitating a dialogue-driven evolution from isolated agents to collective intelligence.
\item Constructing a novel dataset and conducting extensive experiments to validate INMS’s efficacy across diverse open-ended scenarios, demonstrating the its capability to sustain high-quality multi-agent discourse and dynamically build reliable interaction histories.
\end{itemize}

\section{Related Work}

\subsection{Memory Operations}
Equipping Large Language Model (LLM)-based agents with memory mechanisms has gained significant research interest, primarily aiming to enhance conversational consistency, behavioral stability, and experience accumulation. For example, generative agents utilize memory features to store extensive experiential records, facilitating deeper self-understanding \citep{park2023generative}. VOYAGER introduces an evolving skill library that incorporates successful action programs, optimizing task resolution capabilities \citep{wang2023voyager}. Similarly, Ghost in the Minecraft employs a text-based memory system for efficient plan formulation based on historical references \citep{zhu2023ghost}. Memochat adopts a "memorization-retrieval-response" model to sustain coherent, long-range conversations \citep{lu2023memochat}. MemGPT presents a hierarchical memory system tailored for processing extensive textual data and maintaining long-term memory \citep{packer2023memgpt}, while TiM supports continuous memory evolution by storing historical conversational streams \citep{liu2023think}. Reflexion \citep{shinn2023reflexion} and SYNAPSE \citep{zheng2023synapse} further utilize episodic memory buffers and exemplar memory for enhanced decision-making and generalization of successful behaviors, respectively. However, the aforementioned methods primarily focus on enhancing individual agent performance through internal memory operations, without considering interactive memory sharing among multiple agents. In contrast, our proposed INMS framework introduces an interactive memory-sharing mechanism, enabling collective self-enhancement among agents. INMS facilitates memory sharing and mutual learning, fostering collective intelligence beyond individual memory capacities.

\subsection{In-Context Learning}
In-Context Learning (ICL) significantly enhances the problem-solving capabilities of LLMs by integrating few-shot examples directly into prompts \citep{brown2020language, levine2021inductive, zhou2022large, liu2023pre, white2023prompt, gao2024vrsd}. Recent studies illustrate ICL's effectiveness in promoting creative outputs \citep{swanson2021story}, improving logical reasoning through optimized input designs \citep{wiegreffe2021reframing, wu2022ai}, and refining task execution via crowdsourced instructions \citep{mishra2021cross}. Moreover, explicitly highlighting relationships between provided examples and targeted tasks substantially benefits LLM comprehension \citep{lampinen2022can}. Approaches such as Chain-of-Thought (CoT) prompting \citep{wei2022chain} and PAL \citep{gao2023pal} further augment reasoning capabilities by introducing intermediate reasoning steps. Despite these advancements, open-ended question handling remains challenging due to insufficient problem descriptions and the limited availability of high-quality external knowledge bases. Our INMS framework directly addresses these challenges by dynamically generating and sharing high-quality memories among multiple agents, providing rich, diverse reference examples, thereby enhancing agents' performance.

\subsection{Retrieval-Augmented Generation}
Retrieval-Augmented Generation (RAG) methods aim to enhance LLM-generated responses by integrating retrieval techniques such as BM25 \citep{luo2023dr,liu-etal-2022-makes} or SBERT \citep{reimers2019sentence}, improving content accuracy and timeliness \citep{lewis2020retrieval, ram2023context, shi2023replug, gao2026pooling}. Recent studies further boost performance by employing dense retrievers trained via contrastive learning \citep{rubin2021learning}. Additionally, iterative retriever training can enhance retrieval effectiveness \citep{wang2023learning}. Nevertheless, most retrievers in current RAG implementations undergo training only once before deployment, limiting their adaptability to dynamically evolving data. Unlike previous approaches, the INMS framework incorporates a continuously updated retriever, dynamically retrained as new memories are integrated into the memory pool. This approach ensures sustained improvement in retrieval quality and adaptability to evolving data, thereby reducing dependence on external databases and enhancing overall performance.

\section{The Interactive Memory Sharing}
\begin{figure*}[t]
\begin{center}
\includegraphics[width=1\linewidth]{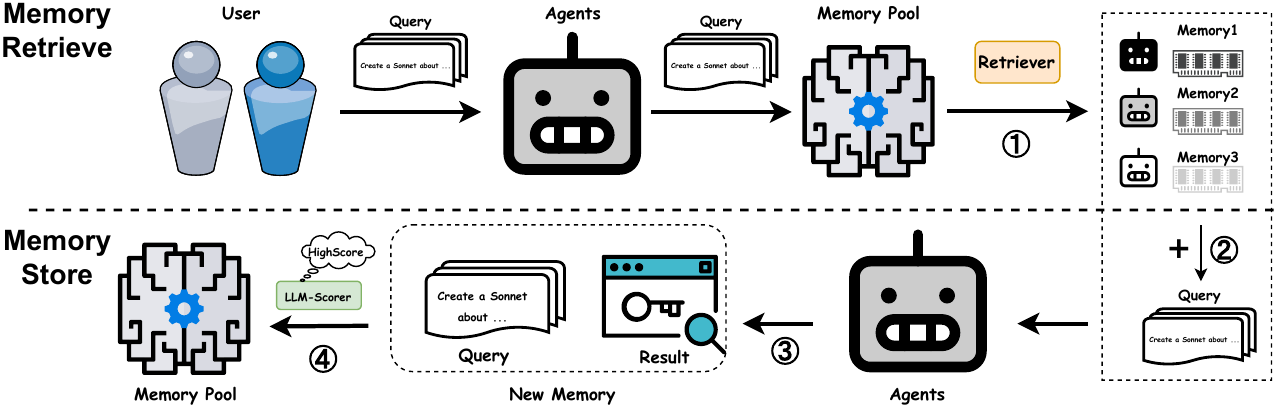}
\caption{The architecture of INMS framework. (1) + (2) The retriever take the original query from agent as the input, retrieve the suitable memories from the memory pool and concatenate them to the query to form the prompt. (3) The Agent (Sonnet) takes the prompt and makes an answer, pack them as (Prompt, Answer). (4) Scorer generates a score according to the designed rubric for (Prompt, Answer), while (Prompt, Answer) pairs with high scores will be added into the Memory Pool. All agents share the same Memory Pool; they can write memories into the pool and retrieve memories from the pool so that they can share memories with each other. For a specific example, please refer to Appendix \ref{specific}.}
\vspace{-5mm}
\label{fig:demo}
\end{center}
\end{figure*}
This section details the Interactive Memory Sharing (INMS) framework, designed to improve agent performance through dynamically shared memories, while preserving agent creativity and versatility. Figure.\ref{fig:demo} illustrates the overall INMS architecture. Agents engage in interactions via a Prompt-Answer (PA) pair, generating results that undergo rigorous evaluation. High-quality PA pairs become shared memories, enriching a common memory pool accessible to all agents. These memories subsequently enhance agent responses and continuously improve the memory retriever. Over successive interactions, the shared memory pool evolves, improving response quality and facilitating a shift from individual to collective intelligence. The core components of INMS are detailed below.

\subsection{Memory Generation and Selection}

A memory is essentially a Prompt-Answer (PA) pair. In some cases, it is permissible for a PA pair to lack a prompt, typically applicable in initial scenarios. These PA pairs are stored in natural language, which serves as the shared memories. These shared memories can be used to improve the response quality of agents. The dynamic expansion of the shared memory pool ensures a continuous influx of new memories, thereby enriching the datasets of agents. In addressing open-ended questions, these shared memories provide agents with a broader perspective and deeper understanding.

After each interaction, the PA pair is scored by a LLM scorer. For each newly generated memory, an LLM scorer will grade it and decide whether to add it to the pool. Before grading, we will establish grading rubrics for each domain, which will be shared among agents within the same domain. To generate these rubrics, we first query LLM several times to obtain various sets of rubrics. Then, we let the LLM evaluate these sets and extract the most reasonable rubrics, synthesizing a complete and useful set. This set of rubrics undergoes a manual review phase to assess the relevance of potential memories to the current focal task and their domain-specific utility, providing additional precision and consideration for the users' specific needs, particularly in specialized application scenarios. The final scoring criteria are constructed by combining individual rubric into a comprehensive criteria. Once this set of rubrics is finalized, every new memory generated will be combined with the corresponding rubric and submitted to the LLM scorer. If the score of PA pair exceeds a preset threshold, the answer and its corresponding prompt are packaged as a useful memory and stored in the memory. During the grading phase, different from the traditional method of directly giving a total score, we prompt LLM to assign a score range for each single rubric in the scoring criteria. Once the score ranges for all rubrics are collected, the final score will be given by the following formula:
\begin{equation}
S_{\text{final}} = \frac{1}{2} \left( \sum_{i=1}^{n} L_i + \sum_{i=1}^{n} H_i \right)
\end{equation}
where \( n \) be the number of rubrics in the scoring criteria. \( L_i \) represents the lowest score in the range for rubric \( i \), and \( H_i \) is the highest score in the range for rubric \( i \).

The different scoring criteria for various domains ensure the specificity of scoring. While the autonomously generated grading criteria by the LLM based on the assumption that the LLM-based agents can better grasp criteria it designed. Therefore, these scoring criteria are established prior to the deployment of the framework to ensure consistency in the LLM's scoring process, thereby guaranteeing fair evaluation of memories from agents. And the manual review phase of rubrics, assessing the relevance of potential memories to the current focal task and their relevance within the domain to ensure their utility, provide additional precision and special consideration to align with specific needs. 

\subsection{Memory Retrieval and Training}
Prior to the deployment of INMS, a small subset of instances, already graded by the same LLM scorer and surpassing the preset threshold, was manually archived within the memory pool as a preliminary step to eliminate potential bias. These instances fulfill a dual purpose: firstly, they provide a diversified array of memories upon which agents may experiment with novel prompts in the face of new queries; secondly, they constitute the preliminary training corpus for our retriever. This foundational training regimen mirrors the methodology by which subsequently archived memories will be assimilated into our model in real time, thereby facilitating the model's ongoing adaptive learning and optimization. During the answering phase, agents retrieves memories from the shared memory pool based on the question with the help of a dense retriever, which are more similar to the target question in terms of cosine similarity. These retrieved memories, combined with question, form a prompt that is submitted to agents, which then generates an answer. The memories extracted from the shared memory are used as context to enhance the quality of the agents' response, a typical ICL method that usually improves the answer quality.
\begin{table*}[ht]
\begin{center}
\resizebox{\textwidth}{!}{
\begin{tabular}{lcccccccccccc}
\toprule
\multicolumn{1}{c}{} & \multicolumn{4}{c}{\bf open-mistral-7b} & \multicolumn{4}{c}{\bf gpt-3.5-turbo} & \multicolumn{4}{c}{\bf gpt-4o} \\
\cmidrule(r){2-5} \cmidrule(lr){6-9} \cmidrule(l){10-13}
\bf Agent & \bf Zero & \bf One & \bf Two & \bf Three & \bf Zero & \bf One & \bf Two & \bf Three & \bf Zero & \bf One & \bf Two & \bf Three \\
\midrule
Limerick   & 0.49 & 0.54 & 0.56 & \textbf{0.59} & 0.50 & 0.56 & 0.76 & \textbf{0.87} & 0.52 & 0.69 & 0.88 & \textbf{0.93}  \\
Wuyanlvshi   & 0.56 & 0.59 & 0.61 & \textbf{0.66} & 0.66 & 0.72 & 0.71 & \textbf{0.72} & 0.73 & 0.75 & 0.75 & \textbf{0.76}  \\
Sonnet   & 0.48 & 0.52 & 0.52 & \textbf{0.52} & 0.50 & 0.53 & 0.53 & \textbf{0.53} & 0.52 & 0.55 & 0.54 & \textbf{0.54} \\
Puzzle   & 0.42 & 0.48 & 0.48 & \textbf{0.50} & 0.47 & 0.51 & 0.52 & \textbf{0.52} & 0.53 & 0.53 & 0.56 & \textbf{0.60} \\
Pun   & 0.32 & 0.35 & 0.36 & \textbf{0.39} & 0.47 & 0.57 & 0.64 & \textbf{0.67} & 0.61 & 0.64 & 0.67 & \textbf{0.70} \\
Riddle   & 0.34 & 0.36 & 0.37 & \textbf{0.37} & 0.40 & 0.42 & 0.48 & \textbf{0.52} & 0.64 & 0.70 & 0.86 & \textbf{0.88} \\
Fitness   & 0.47 & 0.48 & 0.50 & \textbf{0.54} & 0.42 & \textbf{0.57} & 0.52 & 0.52 & 0.46 & 0.61 & 0.65 & \textbf{0.65} \\
Study  & 0.41 & 0.45 & 0.46 & \textbf{0.46} & 0.49 & \textbf{0.56} & 0.53 & 0.51 & 0.55 & 0.60 & 0.63 & \textbf{0.65} \\
Travel   & 0.44 & 0.48 & 0.50 & \textbf{0.53} & 0.47 & 0.54 & 0.54 & \textbf{0.54} & 0.53 & 0.55 & 0.71 & \textbf{0.71} \\
\bottomrule
\end{tabular}
}
\caption{Performance across agents utilizing different amounts shared memories.}
\label{tab:table1}
\end{center}
\end{table*}

\textbf{Memory Train.} Whenever a new PA pair (memory), denoted as \((X,Y)\), is added into the memory pool, it will also be used to train our retriever, which help the retriever to continuously update itself and continuously adapt to new memory. Only PA pairs that pass the memory-quality threshold trigger retriever updates. For each accepted memory, retriever retrieves candidate memories from the existing pool, and the LLM scorer provides relative-reference labels for updating the retriever. The updates can be processed asynchronously in batches without blocking response generation. Previously stored memories remain available and continue to serve as candidate examples during later updates. Based on the new generated memory \((X,Y)\), the classical method BM25 ascertain the most pertinent top-$n$ candidate pairs $\{(x_i, y_i)\}_{i=1}^n$, sourced from the diverse and extensive memory pool, denoted as $C$. Each candidate within \(C\) will undergo a evaluation process utilizing the comprehensive scoring capabilities of LLM. The scoring mechanism employed is defined as following:
\begin{equation}
p(x_i, y_i) =  \text{P}(\neg Y \mid (x_i, y_i),\, X), i \in \{1,..., n\}
\end{equation}

\begin{table*}[ht]
\centering
\resizebox{\textwidth}{!}{
\begin{tabular}{l*{9}{cc}}
\toprule
\multirow{3}{*}{\textbf{Methods}}
  & \multicolumn{6}{c}{\textbf{Literary}} & \multicolumn{6}{c}{\textbf{Logic}} & \multicolumn{6}{c}{\textbf{Plan}} \\
\cmidrule(lr){2-7}\cmidrule(lr){8-13}\cmidrule(lr){14-19}
  & \multicolumn{2}{c}{\textbf{Limerick}}
  & \multicolumn{2}{c}{\textbf{Wuyanlvshi}}
  & \multicolumn{2}{c}{\textbf{Sonnet}}
  & \multicolumn{2}{c}{\textbf{Puzzle}}
  & \multicolumn{2}{c}{\textbf{Pun}}
  & \multicolumn{2}{c}{\textbf{Riddle}}
  & \multicolumn{2}{c}{\textbf{Fitness}}
  & \multicolumn{2}{c}{\textbf{Study}}
  & \multicolumn{2}{c}{\textbf{Travel}} \\
\cmidrule(lr){2-3}\cmidrule(lr){4-5}\cmidrule(lr){6-7}
\cmidrule(lr){8-9}\cmidrule(lr){10-11}\cmidrule(lr){12-13}
\cmidrule(lr){14-15}\cmidrule(lr){16-17}\cmidrule(lr){18-19}
  & \textbf{F1} & \textbf{LLM-J}
  & \textbf{F1} & \textbf{LLM-J}
  & \textbf{F1} & \textbf{LLM-J}
  & \textbf{F1} & \textbf{LLM-J}
  & \textbf{F1} & \textbf{LLM-J}
  & \textbf{F1} & \textbf{LLM-J}
  & \textbf{F1} & \textbf{LLM-J}
  & \textbf{F1} & \textbf{LLM-J}
  & \textbf{F1} & \textbf{LLM-J} \\
\midrule
Random
  & 0.21 & 0.15 & 0.00 & 0.40 & 0.16 & 0.50
  & 0.17 & 0.14 & 0.25 & 0.15 & 0.28 & 0.38
  & 0.14 & 0.15 & 0.09 & 0.03 & 0.11 & 0.02 \\
BM25
  & 0.44 & 0.90 & 0.00 & 0.50 & 0.18 & 0.55
  & 0.19 & 0.22 & 0.41 & 0.20 & 0.65 & 0.60
  & 0.20 & 0.30 & 0.13 & 0.08 & 0.18 & 0.08 \\
Contriever
  & 0.43 & 0.92 & 0.00 & 0.49 & 0.18 & 0.53
  & 0.18 & 0.25 & 0.39 & 0.23 & 0.63 & 0.55
  & 0.21 & 0.38 & 0.12 & 0.05 & 0.16 & 0.04 \\
SBERT
  & 0.43  & 0.88 & 0.00 & 0.46 & 0.17 & 0.58
  & 0.16 & 0.27 & 0.40 & 0.18 & 0.65 & 0.58
  & 0.19 & 0.23 & 0.12 & 0.04 & 0.15 & 0.05 \\
TAS\mbox{-}B
  & 0.44 & 0.93 & 0.00 & 0.52 & 0.14 & 0.52
  & 0.14 & 0.21 & 0.35 & 0.15 & 0.63 & 0.50
  & 0.19 & 0.23 & 0.13 & 0.15 & 0.15 & 0.07 \\
SimCSE
  & 0.41 & 0.93 & 0.00 & 0.47 & 0.17 & 0.58
  & 0.15 & 0.19 & 0.40 & 0.18 & 0.64 & 0.58
  & 0.21 & 0.32 & 0.11 & 0.09 & 0.19 & 0.12 \\
\cem{INMS}
  & \cem{\textbf{0.51}} & \cem{\textbf{0.97}} & \cem{\textbf{0.00}} & \cem{\textbf{0.55}} & \cem{\textbf{0.24}} & \cem{\textbf{0.67}}
  & \cem{\textbf{0.18}} & \cem{\textbf{0.31}} & \cem{\textbf{0.38}} & \cem{\textbf{0.24}} & \cem{\textbf{0.69}} & \cem{\textbf{0.71}}
  & \cem{\textbf{0.23}} & \cem{\textbf{0.34}} & \cem{\textbf{0.13}} & \cem{\textbf{0.11}} & \cem{\textbf{0.21}} & \cem{\textbf{0.15}} \\
\bottomrule
\end{tabular}}
\caption{Comparison of INMS with baselines under both F1 and LLM Judge (LLM-J) metrics.}
\label{tab:mainresults}
\end{table*}

This equation seeks to determine, given a input-output pair \((x_i, y_i)\) in \(C\) as a condition, the probability that the response generated for the input in the new memory contradicts the output in the new memory. This grading part serves as a preparatory step for the subsequent labeling of each candidate example. It is noteworthy that making $\neg Y$ as the result part is trying to make sure that the memory that the retriever gets from agents is of reference value, but it does not have to be the most relevant to the current question, which means that it can help the current agent to learn from new examples. This approach diverges from a simplistic reliance on \(Y\) as the outcome, which tends to restrict the retrieval process to memory previously stored by agents. 

Within the defined set \(C = \{(x_i, y_i)\}_{i=1}^n\), each candidate now is ascribed a score. We sort them from the lowest to the highest score and we select $v$ memory in total to label. The top $\frac{v}{2}$ candidates (lowest score) in \(C\) are identified as being the pair with the reference value to \((X, Y)\) and accordingly, their labels are set to positive. Conversely, the bottom $\frac{v}{2}$ candidates are deemed as the least reference value to \((X, Y)\), and their labels are thus designated as negative. Those labeled data will be used to minimize the following function:
\begin{equation}
\begin{aligned}
    \text{loss}(x, y) = -\frac{1}{v} \sum_{i=1}^{v}[ y_i \cdot \log(\frac{1}{1 + e^{-x_i}}) + \\
    (1 - y_i) \cdot \log(1 - \frac{1}{1 + e^{-x_i}})]
\end{aligned}
\end{equation}

\subsection{Interactive Learning}
INMS employs an interactive learning strategy to rapidly expand and update the memory pool, particularly addressing initial memory scarcity. Initially, a small number of PA pairs (as few as 100 or even one) are stored in the memory pool. Agents iteratively generate prompts based on provided answers, subsequently answering these prompts to create additional PA pairs. High-quality pairs identified through scoring are added to the memory pool, accelerating memory growth and agent self-enhancement. This interactive approach also underpins the construction of our experimental datasets, allowing continuous evaluation and refinement of agent performance.

\section{Experiments}
\subsection{Baseline}
We evaluate INMS from two perspectives. First, we analyze generator sensitivity by testing multiple LLMs while varying the number of retrieved memories $k\in\{0,1,2,3\}$; when $k=0$, the model receives only the task prompt, and as $k$ increases, the top-$k$ retrieved memories are included as contextual examples. Second, we compare INMS with retrieval baselines under a fixed generator setting. The baselines include \textbf{Random}, which samples memories uniformly; \textbf{BM25} \citep{article}, a sparse lexical retriever; \textbf{Contriever} \citep{izacard2021unsupervised}, an unsupervised dense retriever trained with momentum contrast; \textbf{SBERT} \citep{reimers2019sentence} using the all-mpnet-base-v2 model; \textbf{TAS-B} \citep{hofstatter2021efficiently}, a topic-aware dense retriever trained with balanced margin sampling; and \textbf{SimCSE} \citep{gao2021simcse}, a contrastive sentence-embedding framework. 

\subsection{Implementation Details}
We aim to assess the efficacy of the INMS in processing open-ended questions across three domains: Literary Creation, Unconventional Logic Problem-solving, and Plan Generation. Separate memory pools were allocated for each of the three domains because these domains are unrelated and have no overlapping content. Within the Literary Creation domain, three agents were assigned the tasks of generating Wuyanlvshi (a classical form of Chinese poetry), Limericks, and Sonnets, respectively. In the Logic Problem-solving domain, three agents were tasked with solving Puzzles, Riddles, and Puns separately. For the Plan Generation domain, agents were employed to create Study Plans, Travel Plans, and Fitness Plans individually. Each of these nine agents were associated with a corresponding dataset used for evaluation. Across all nine datasets, comprising a total of 1000 instances (details provided in Appendix \ref{datasets}), we partitioned the data within each dataset into three subsets: 20\% was allocated for constructing the initial memory pool, 40\% was used to extract queries that were then input back into the agents to generate memory, and the remaining 40\% was reserved as the test set. The three subsets are mutually exclusive. Test instances are used only for final evaluation and are excluded from initial memory construction, interaction-based memory generation, and retriever updates. For our scoring LLM, we use gpt-4o. As the backbones of our agents, we first consider three LLMs: two close-source LLMs (gpt-3.5-turbo and gpt-4o) \citep{achiam2023gpt} and one open-source LLM (open-mistral-7b) \citep{Jiang2023Mistral7}. We use the BERTScore \citep{bert-score} and F1 score as our metric to measure the performance of each agent. Besides, we adopt an LLM Judge (LLM-J)evaluation, where an LLM determines whether a model’s answer is semantically consistent with the ground truth, providing a reliable alternative to surface-based metrics \citep{gu2024survey}. The threshold for selecting PA pairs is fixed at 81 as a global quality cutoff obtained from the average scorer outputs across the dataset. The same threshold is applied unchanged across all experimental settings and is not selected according to downstream test performance. We report the experiment results with extra metrics in the Appendix \ref{extra}.

It is worth noting that, before the experiment, none of the agents have a suitable database for reference. While after the interactive learning stage, a continuously expanding memory pool with high quality memories is successfully be a database for agents to refer. The INMS framework help agents get rid of the dependence on external databases, and agents can interactively expand the memory pool without taking a lot of effort. 
\subsection{Main Results}
\begin{figure}[t]
\begin{center}
\includegraphics[width=0.8\columnwidth]{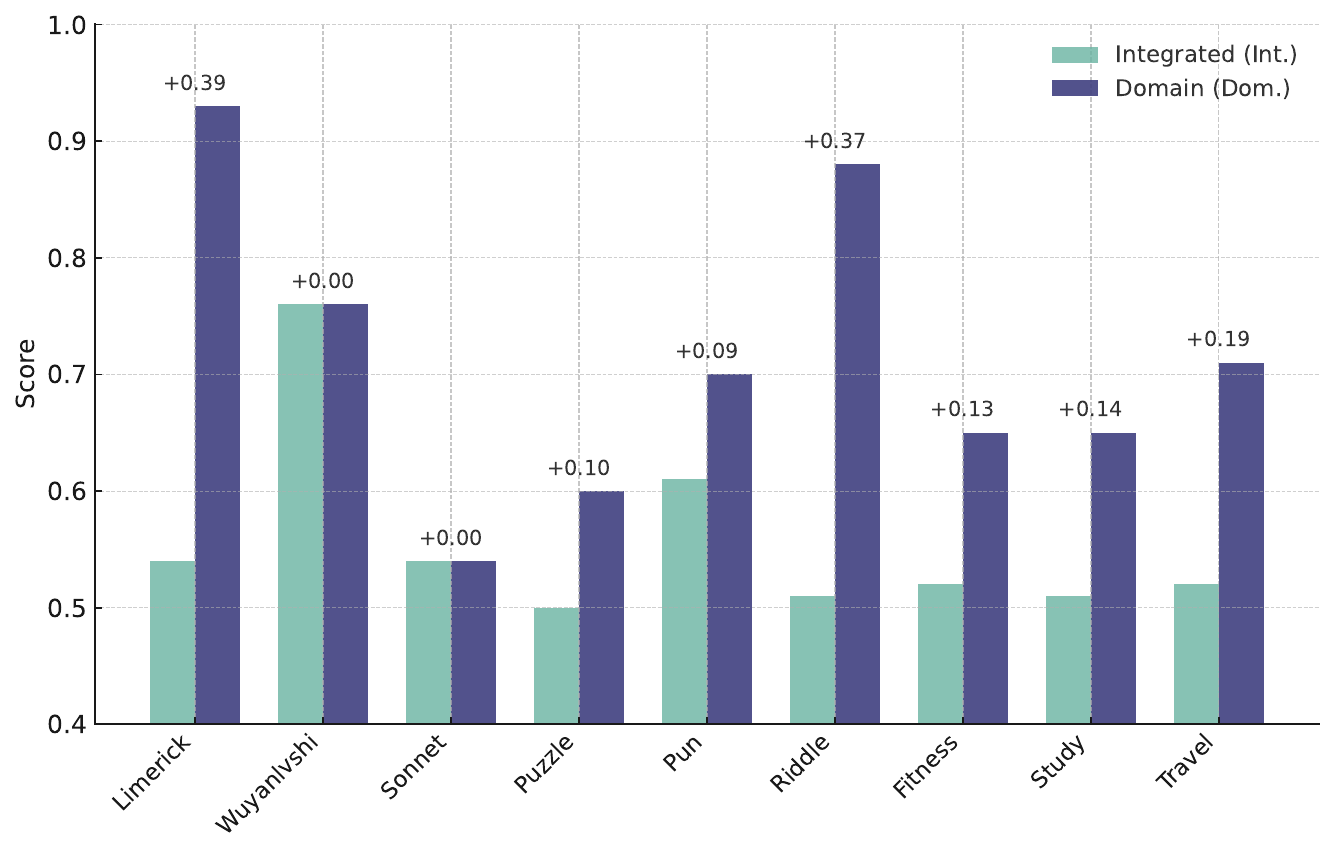}
\caption{Domain Pool vs Integrated Pool Performance}
\vspace{-7mm}
\label{fig:DvsI}
\end{center}
\end{figure}
For each agent, we first tested them with using the same backbone and the metric BertScore, that is, in each domain, all memory was generated by agents utilizing the same Large Language Model, and in subsequent task execution, the memory generated by the previous execution of other agents could be used. Table.\ref{tab:table1} shows the result of each agent. We can observe that, for all agents among all the tasks, compare to no use of the shared memories, the performance of all the agents has been improved. This suggests that the shareable memories from other tasks can help agents get desired answers, rather than interfering with the agents' learning ability. These results show that the complete INMS pipeline consistently improves agent performance in the shared-memory setting. Further separating the contributions of self-generated and cross-agent memories provides an interesting direction for future analysis. And employing more shared memories leads to further performance enhancements across nearly all agents. This improvement underscores the effectiveness of shared memories, attributable to the constantly updated retriever's ability to adjust as the memory pool expands. Consequently, retrievers can consistently retrieve the most relevant PA pairs for each query. Besides, when using the same number of shared memories, the closed-source LLM demonstrates superior performance compared to the open-source LLM, likely due to its enhanced understanding and reasoning capabilities. Additionally, since using three shared memories yields the best performance, all subsequent experiments utilize three shared memories during testing and gpt-4o as the backbone.

\begin{figure}[t]
\begin{center}
\includegraphics[width=0.8\columnwidth]{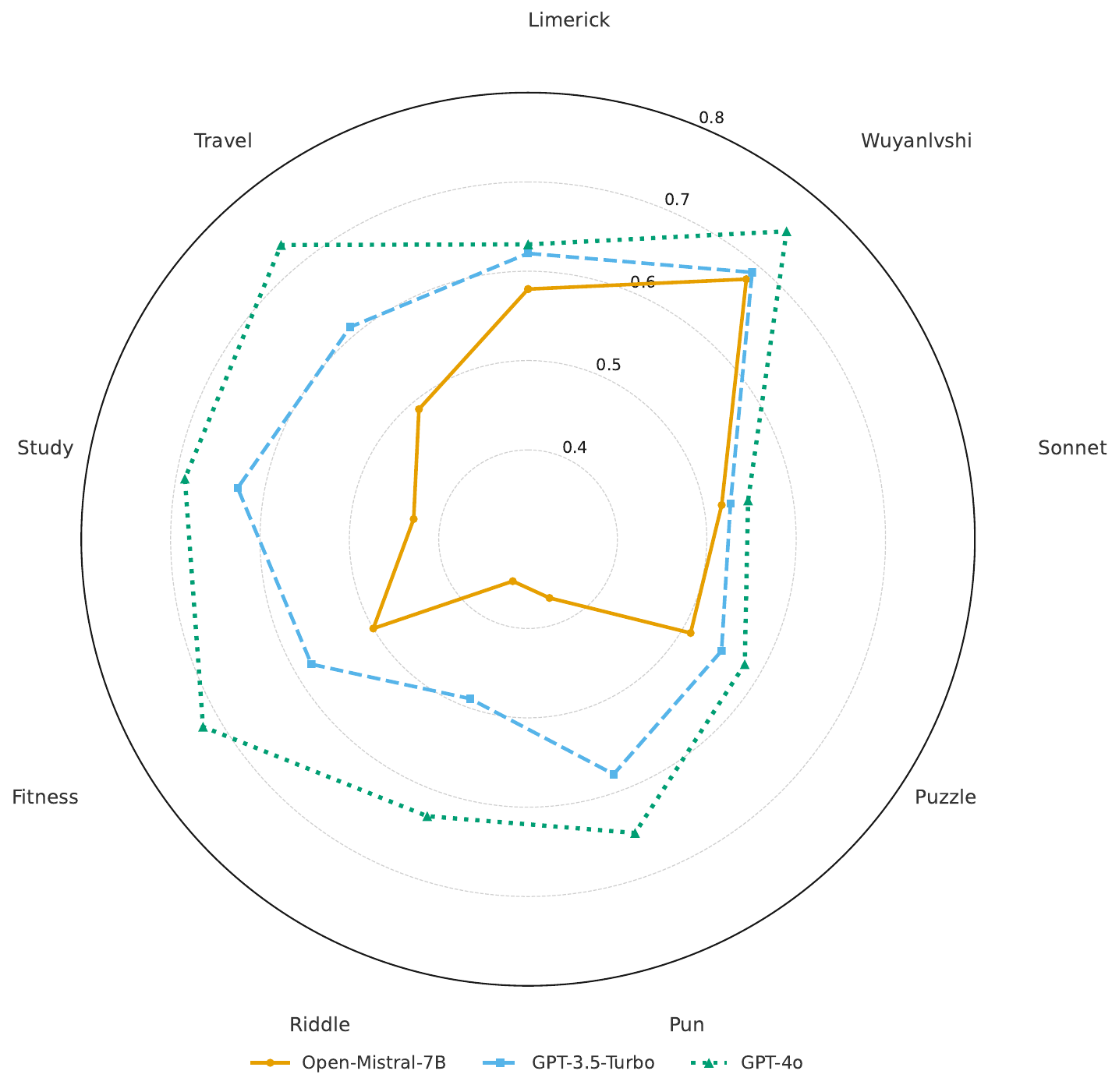}
\caption{Performance across agents by utilizing heterologous shared memories.}
\label{fig:heter}
\end{center}
\end{figure}
As shown in Table.\ref{tab:mainresults}, INMS consistently surpasses both sparse and dense retrieval baselines. Traditional methods such as BM25 and Random exhibit limited ability to generalize across diverse query types, while contrastive-based dense retrievers achieve moderate gains but remain constrained by static embeddings that fail to adapt to evolving memory contexts. In contrast, INMS demonstrates robust improvements in both F1 and LLM-judge metrics across literary, logical, and planning tasks, indicating that interactive memory sharing effectively enhances retrieval relevance and generation quality. The adaptive retriever plays a key role in this improvement by continuously aligning the embedding space with newly admitted memories, allowing agents to draw on more contextually useful information. Moreover, INMS achieves the most balanced performance across creative and reasoning-oriented domains, highlighting its capacity to integrate collective memory without overfitting to any single task type.
\subsection{Does Domain-Specific Memory Outperform Integrated Memory?}
\begin{figure*}[t]
\begin{center}
\includegraphics[width=1\linewidth]{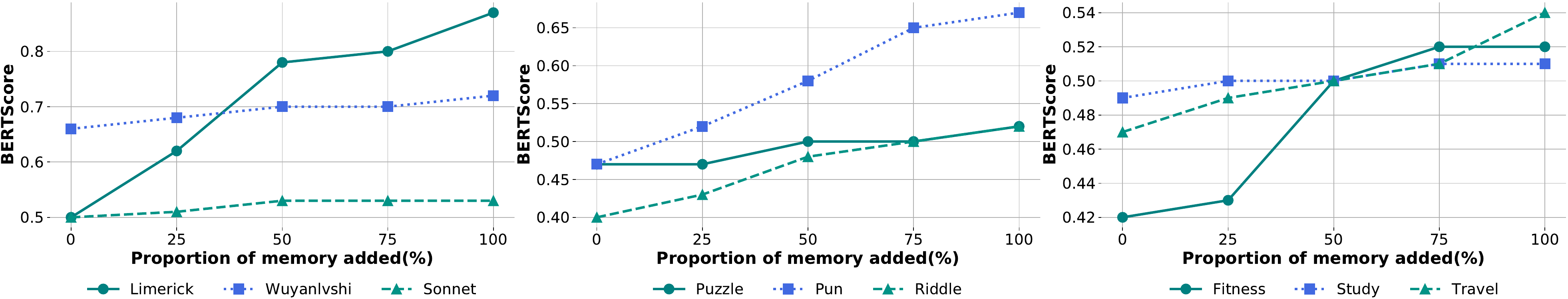}
\caption{Evaluating agents' performance on accumulation of memories in five phases.}
\vspace{-5mm}
\label{fig:figure33}
\end{center}
\end{figure*}

\begin{figure*}[t]
\begin{center}
\includegraphics[width=1\linewidth]{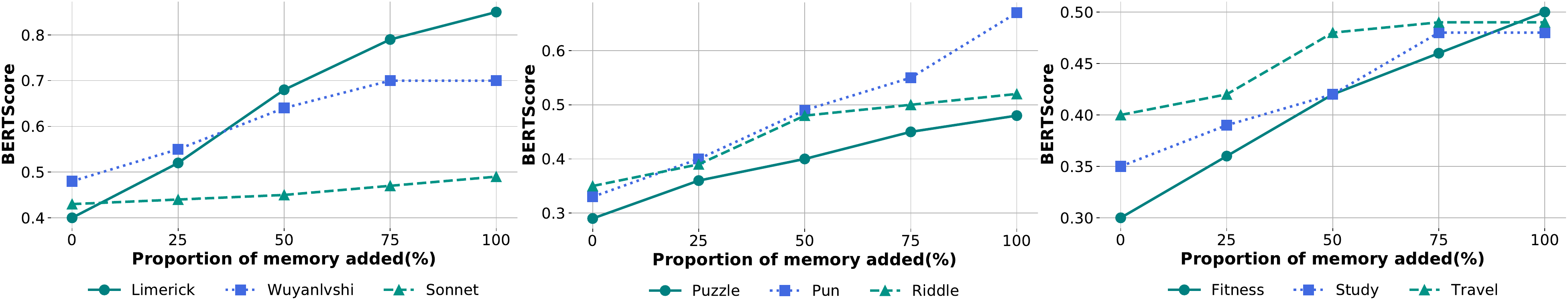}
\caption{Evaluating agents' performance on biased initial memory pool in five phases.}
\vspace{-7mm}
\label{fig:figure4}
\end{center}
\end{figure*}
Besides, since memories from agents in other domains may help agents better understand queries from different angles and enrich the diversity of the memory pool, we constructed an additional pool, the integrated pool, which combines all shareable memories from all agents across all domains into a single pool. However, as shown in Figure.\ref{fig:DvsI}, although the integrated pool can enhance the diversity of shared memories, the domain-specific pool more effectively enables agents to produce reliable answers, regardless of the LLM used by the agents.

\subsection{Can Shareable Memories Benefit Agents Utilizing Different LLM Backbones?}

To verify that shareable memories generated by agents utilizing different LLMs during task execution can still aid current agent in processing queries, we deployed agents based on three distinct LLMs to execute an equal number of queries in each domain, thereby expanding the memory pool. Also, those potential shareable memories only after being scored by the LLM scorer will be decided whether they can be added to the memory pool. The results in Figure.\ref{fig:heter} show that, compared to the non-use of shared memories in Table.\ref{tab:table1}, the cross LLMs shareable memories can still boost the performance for all the agents in answering the open-ended questions, which means that those heterologous shared memories still be useful for agents. Although the performance of agents do not always appear the same trend compared to the use of same amount shared memories from itself, that is, either rising or falling, they all ultimately improve agents’ answers, but the degree varies.

\subsection{Dynamics of Interaction Accumulation and Mitigating the Echo Chamber Effect}
Next, we investigate whether an extensive accumulation of interaction history impedes the agents' conversational quality, and whether an initial "echo chamber", a shared memory pool dominated by biased early interactions, can be effectively dismantled as the system evolves. This recovery is facilitated by the LLM-scorer (acting as a dialogue moderator) and the continuously updated retriever. We measure agent performance across five phases of interaction accumulation—0, 25\%, 50\%, 75\%, 100\% expansion of the shared memory pool.
To simulate an initial echo chamber, we deliberately constructed a biased starting pool for each domain, where 75\% of the prompt-answer (PA) pairs exhibit significant bias. The LLM scorer evaluated these biased communicative acts, with half receiving scores near 0 and the other half near 50. We then tracked the agents' performance through the same five evolutionary phases.
The results presented in Figure.\ref{fig:figure33} highlight that as high-quality interactive memories are continuously infused into the shared pool, the collective performance of the agents steadily improves. For several tasks, performance plateaus in the later stages. We hypothesize that this stabilization occurs because the shared conversational ground has reached a temporary saturation of optimal examples; further expansion or diversification of the interaction topology may be required to break this stagnation.
Crucially, Figure.\ref{fig:figure4} demonstrates the system's resilience against the echo chamber effect. Although the initially biased pool heavily interfered with the agents' responses during early interactions (as they repeatedly retrieved and mirrored flawed examples), the continuous influx of new, dynamically filtered memories steadily diluted this bias. Consequently, the agents' performance rebounded, ultimately approaching the peak scores observed in the unbiased scenario (Figure.\ref{fig:figure33}). These findings show that the LLM scorer and dynamic retriever can progressively reduce the influence of initially biased memories as higher-quality memories accumulate. The scoring mechanism moderates newly admitted memories, while the updated retriever increasingly prioritizes more useful memories for subsequent queries.

\section{Conclusions}
This study introduces INMS, a novel framework modeling real-time memory sharing among agents as asynchronous dialogue via continuous storage and retrieval. This dynamically evolving shared memory acts as a conversational ground, enhancing agents' ability to collaboratively understand nuances and generate high-quality responses. Even for uncommon problems, INMS rapidly builds a robust interaction history. Experiments demonstrate its effectiveness in modeling multi-agent communication, successfully overcoming early-stage "echo chamber" biases as the pool stabilizes. Furthermore, our newly constructed dataset fills critical gaps in open-ended generation benchmarks. Ultimately, INMS paves the way for agents to evolve from isolated solvers into a collectively intelligent society driven by continuous dialogue and knowledge exchange.

\section*{Limitations}
Our INMS framework shows encouraging performance, but there remain several directions worth pursuing. First, this study emphasizes validating the core idea of Interactive Memory Sharing rather than exhaustive systems tuning. To keep the setup clear and reproducible, we evaluate on representative domains and defer broader generalization to later work. Second, although the framework applies memory filtering, the resulting organization can still be shaped by the capabilities of the underlying language models. Finally, our current prototype targets text-only interactions; extending it to multimodal inputs, e.g., images or audio, could supply richer context and is a natural avenue for future research.

\bibliography{custom}
\appendix

\section*{Appendix}
\label{sec:appendix}
\section{Rubrics and Prompt for scoring Memory}
In order to judge whether a memory can be added into the memory pool, we set three scoring rubrics for three domains respectively. For Single Pool, we set up a set of rubrics from a global perspective.

\subsection{Rubrics for domain - Literary Creation}
\vspace{4mm}
\hrule
\vspace{2mm}
\textbf{General Evaluation Criteria (Total: 100)}\\
\textbf{Criteria: Literary Quality} \\
\textbf{Score Range: 0-5} \\
\textbf{Description:} Assesses creativity, use of language, and emotional impact. High-quality examples should demonstrate mastery of language and evoke a strong reader response.\\
\textbf{Criteria: Authenticity} \\
\textbf{Score Range: 0-10} \\
\textbf{Description:} Evaluates adherence to the form's traditional standards, including structure, rhythm, and themes. High scores indicate that the poem respects genre conventions creatively.\\
\textbf{Criteria: Clarity and Cohesion} \\
\textbf{Score Range: 0-10} \\
\textbf{Description:} Considers the poem's clarity of expression and the cohesion of its parts. A high score indicates that the poem communicates effectively and its elements are well integrated.\\
\textbf{Criteria: Innovativeness} \\
\textbf{Score Range: 0-5} \\
\textbf{Description:} Rewards originality in theme, structure, or language use. High scores reflect a notable degree of creativity and the introduction of novel ideas or techniques.\\
\textbf{Criteria: Educational Value} \\
\textbf{Score Range: 0-10} \\
\textbf{Description:} Assesses the example's potential to teach about poetic forms, literary devices, and thematic exploration. High-scoring examples are rich in analyzable and teachable elements.\\
\textbf{Criteria: Metric Precision} \\
\textbf{Score Range: 0-10} \\
\textbf{Description:} Evaluates the adherence to the five-syllable structure per line, including rhythm and flow, emphasizing the importance of metric accuracy.\\
\textbf{Criteria: Imagery and Symbolism} \\
\textbf{Score Range: 0-10} \\
\textbf{Description:} Assesses the effectiveness of imagery and symbolism in conveying the poem's themes, highlighting the depth and sophistication of language use.\\
\textbf{Criteria: Humor and Wit} \\
\textbf{Score Range: 0-10} \\
\textbf{Description:} Rates the poem's humor, wit, and wordplay. High scores reflect effective use of language to entertain and amuse.\\
\textbf{Criteria: Rhyme Scheme Adherence} \\
\textbf{Score Range: 0-10} \\
\textbf{Description:} Assesses the AABBA rhyme scheme's quality and creativity, including how well the rhymes enhance the humor and effectiveness of the poem.\\
\textbf{Criteria: Structural Integrity} \\
\textbf{Score Range: 0-10} \\
\textbf{Description:} Evaluates adherence to sonnet structure, including rhyme scheme and division into octaves/sestets or quatrains/couplet, stressing formal precision.\\
\textbf{Criteria: Thematic Development} \\
\textbf{Score Range: 0-10} \\
\textbf{Description:} Looks at theme or argument development, especially through the volta, reflecting the poem's ability to engage with complex ideas persuasively.\\
\hrule

\subsection{Rubrics for domain - Unconventional Logic Problem-solving}
\vspace{4mm}
\hrule
\vspace{2mm}
\textbf{Clarity and Understandability (20 points)}\\
Question Clarity (10 points): The question should be clearly stated, without ambiguity, and understandable without requiring additional context.\\
Answer Clarity (10 points): The answer should be directly related to the question, clear, and easily understandable.\\
\textbf{Creativity and Originality (30 points)}\\
Question Creativity (15 points): The question should demonstrate creativity, originality, and should not be a common or easily found problem.\\
Answer Creativity (15 points): The answer should be innovative and not just a straightforward or commonly known response. It should also add a layer of depth or a surprising twist to the question.\\
\textbf{Logical Consistency and Correctness (20 points)}\\
Logical Consistency (10 points): The question and answer together should form a logically consistent pair where the answer correctly follows from the question.\\
Correctness (10 points): The answer must be factually correct and provide a true solution or conclusion to the puzzle, riddle, or pun presented in the question.\\
\textbf{Relevance and Engagement (20 points)}\\
Relevance (10 points): The question and answer should be relevant to the domain of Logic Problems, demonstrating an understanding of puzzles, riddles, or puns.\\
Engagement (10 points): The pair should be engaging and interesting, capable of capturing attention and sparking curiosity or amusement.\\
\textbf{Difficulty Level (10 points)}\\
The difficulty level of the question should be appropriate for the intended audience. It should neither be too easy to solve without any thought nor too difficult to be practically unsolvable. This criterion requires a balanced approach to ensure the content is intellectually stimulating but accessible.\\
\hrule

\subsection{Rubrics for domain - Plan Generation}
\vspace{4mm}
\hrule
\vspace{2mm}
\textbf{Specificity and Detail (20 points)}\\
Question Specificity (10 points): The question should be specific, providing enough detail to guide the generation of a relevant and tailored plan.\\
Plan Detail (10 points): The plan should include specific activities, steps, or recommendations that are clearly defined and actionable.\\
\textbf{Feasibility and Practicality (20 points)}\\
Plan Feasibility (20 points): The plan should be realistic and practical, considering available resources (time, money, equipment) and constraints. It should propose actions that can be realistically implemented by the user.\\
\textbf{Comprehensiveness and Scope (20 points)}\\
Coverage of Key Components (20 points): The plan should comprehensively address all relevant aspects of the goal. For a study plan, this might include study sessions, breaks, and topics covered; for a fitness plan, workouts, rest days, and nutrition; and for a travel plan, transportation, accommodations, and activities.\\
\textbf{Personalization and Relevance (20 points)}\\
Alignment with User Needs and Preferences (20 points): The plan should reflect an understanding of the user's specific needs, preferences, goals, and limitations. It should feel customized and directly applicable to the user, rather than being a generic template.\\
\textbf{Plan Clarity (10 points)}: The plan should be articulated in a clear, organized, and easy-to-follow manner. It should avoid jargon or overly complex language, making it accessible to the user.\\
\textbf{Rationale Clarity (10 points)}: The plan should include clear reasoning or justification for the recommendations made, helping the user understand why specific actions or steps are suggested.\\
\hrule

\subsection{Rubrics for Single Pool}
\vspace{4mm}
\hrule
\vspace{2mm}
\textbf{Accuracy (25 Points)}\\
25 points: The output is entirely accurate, with no factual errors or inaccuracies.\\
15-24 points: The output is mostly accurate, with minor errors that do not significantly impact the overall understanding.\\
5-14 points: The output contains several inaccuracies that could lead to misunderstandings.\\
0-4 points: The output is largely inaccurate, misleading, or irrelevant.\\
\textbf{Relevance (20 Points)}\\
20 points: The output is highly relevant to the input question, directly addressing the query without diverging from the topic.\\
10-19 points: The output is relevant but includes some unnecessary or slightly off-topic information.\\
1-9 points: The output partially addresses the question but is significantly off-topic or tangential.\\
0 points: The output is completely irrelevant to the input question.\\
\textbf{Completeness (20 Points)}\\
20 points: The output provides a complete answer to the question, covering all essential aspects implied or directly asked.\\
10-19 points: The output covers most of the necessary information but lacks one or two minor details or aspects.\\
1-9 points: The output provides a partial answer, missing significant portions of the information needed to fully answer the question.\\
0 points: The output fails to provide any meaningful answer to the question.\\
\textbf{Clarity and Coherence (20 Points)}\\
20 points: The output is exceptionally clear and well-structured, making it easy to follow and understand.\\
10-19 points: The output is clear but may have minor issues with structure or coherence that slightly hinder understanding.\\
1-9 points: The output has significant clarity or coherence issues, making it difficult to understand without effort.\\
0 points: The output is incoherent or so poorly structured that it is unintelligible.\\
\textbf{Creativity and Insight (15 Points)}\\
15 points: The output demonstrates high levels of creativity or provides insights that add substantial value beyond the explicit question.\\
8-14 points: The output shows some creativity or insights but to a lesser extent, offering added value to the answer.\\
1-7 points: The output is standard, with minimal to no creativity or insightful additions.\\
0 points: The output is entirely generic, with no attempt at creativity or providing additional insights.\\
\hrule

\subsection{Prompt for scoring Memory}
For scoring a memory, the following instructions were provided to LLM:
\begin{mdframed}
\small
\texttt{Here is the rubrics for grading an (prompt, answer) pair-"grading\_rubric". According to the rubrics, give me a final score of the pair. For the final score, first, for each rubric, according to its standard, give a score range for the current pair, such as 3-6 (3 is the lowest score and 6 is the highest score). After scoring all the rubrics, add up the lowest and highest scores. Then give the average of the two total scores. This score will be the final score. The final score should be a number between 0-100 }
\end{mdframed}

\section{Datasets} \label{datasets}
For all those nine agents, they have their own dataset to evaluate. There are 1000 items in total for all datasets. In each datasets, the item contains a question and a answer. For each dataset, 20\% will be taken randomly as a small subset of instances which was manually archived
within the memory pool before the experiments. And 40\% of the dataset, we will only capture the problem part and use it to generate real-time memory in agents. As the rest 40\%, they will be used as the test set.

For the tasks in the domain Literary Creation, in their datasets, the question part is a description, and the answer part is like a poetry which fullfills the requirements in the description, the question and answer in the dataset are like: 
\begin{mdframed}
\small
\texttt{Tell me about a star that twinkles with a secret that shining brightly, mysteries to decrypt? - There once was a star that twinkles with a secret; With a tale both old and new; shining brightly, mysteries to decrypt; A legend through and through; In a world where magic grew.}
\end{mdframed}

For the tasks in the domain Unconventional Logic Problem-solving  in their datasets, the question part is a logic problem, and the answer part is a reasonable solution of that problem, the question and answer in the dataset are like: 
\begin{mdframed}
\small
\texttt{Why is it better to have round manhole covers than square ones? - A square manhole cover can be turned and dropped down the diagonal of the manhole. A round manhole cover cannot be dropped down the manhole. So for safety reasons, all manhole covers should be round}
\end{mdframed}

For the tasks in the domain Plan Generation, in their datasets, the question part is a problem about how to achieve something, and the answer part is a executable plan, the question and answer in the dataset are like: 
\begin{mdframed}
\small
\texttt{Start learning Python for data analysis? - Learn basics, then focus on data libraries like pandas, and practice on projects.}
\end{mdframed}

Specifically, the dataset of the Puzzle comes from \citet{liu2024agentbench}, and we expanded its size from 70 to 100. The new puzzles are based on some classic problems on the Internet. For other datasets, the standard "answer" was systematically harvested from the internet and subsequently input to the LLM to get the corresponding "question" part (through interactive learning). In particular, the selection of Wuyanlvshi—renowned and historically significant poems within Chinese literature—were meticulously chosen for their fame and widespread recognition. Furthermore, the sonnets incorporated into our study originate from the ”quarto” collection authored by Shakespeare in 1609. We use the following similar format of prompt to generate "question" part: 
\begin{mdframed}
\small
\texttt{Here is a sonnet -> "From fairest creatures we desire increase, That thereby beauty’s rose might never die, But as the riper should by time decrease, His tender heir mught bear his memeory: But thou, contracted to thine own bright eyes, Feed’st thy light’st flame with self-substantial fuel, Making a famine where abundance lies, Thyself thy foe, to thy sweet self too cruel. Thou that art now the world’s fresh ornament, And only herald to the gaudy spring, Within thine own bud buriest thy content And, tender churl, makest waste in niggarding. Pity the world, or else this glutton be, To eat the world’s due, by the grave and thee." According to this, can you give me a problem of it(like make a sonnet followed with some description), so the possible answer can be this sonnet}
\end{mdframed}.
The output as a "question" from the LLM is like:
\begin{mdframed}
\small
\texttt{Craft a sonnet that explores the tension between personal vanity and the broader imperative to contribute to the continuation of beauty and life. Your poem should address how an individual’s focus on their own allure can lead to a metaphorical barrenness, despite the inherent capacity for generational legacy. Conclude with a reflection on the choice between selfishness and altruism, using nature as a metaphor for this existential dilemma.}
\end{mdframed}.

\section{Extra Results} \label{extra}
Table.\ref{tab:table444444444}, Table.\ref{tab:tablettt} and Table.\ref{tab:table3} give the extra results of INMS.
\begin{table*}[!t]
\centering
\resizebox{\textwidth}{!}{
\begin{tabular}{lcccccccc}
\toprule
& \multicolumn{2}{c}{\bf Zero} & \multicolumn{2}{c}{\bf One} & \multicolumn{2}{c}{\bf Two} & \multicolumn{2}{c}{\bf Three} \\
\cmidrule(r){2-3} \cmidrule(lr){4-5} \cmidrule(lr){6-7} \cmidrule(l){8-9}
\bf Agent & \bf ROUGE-2 & \bf ROUGE-L & \bf ROUGE-2 & \bf ROUGE-L & \bf ROUGE-2 & \bf ROUGE-L & \bf ROUGE-2 & \bf ROUGE-L \\
\midrule
Limerick     & 0.06 & 0.15 & 0.25 & 0.37 & 0.44 & 0.52 & 0.75 & 0.77 \\
Wuyanlvshi   & 0.00 & 0.00 & 0.00 & 0.00 & 0.00 & 0.00 & 0.00 & 0.00 \\
Sonnet       & 0.02 & 0.14 & 0.02 & 0.13 & 0.10 & 0.15 & 0.10 & 0.15 \\
Puzzle       & 0.07 & 0.19 & 0.09 & 0.21 & 0.09 & 0.25 & 0.09 & 0.26 \\
Pun          & 0.27 & 0.43 & 0.20 & 0.35 & 0.30 & 0.43 & 0.24 & 0.37 \\
Riddle       & 0.71 & 0.80 & 0.32 & 0.48 & 0.44 & 0.56 & 0.62 & 0.75 \\
Fitness      & 0.02 & 0.06 & 0.04 & 0.15 & 0.06 & 0.18 & 0.07 & 0.19 \\
Study        & 0.01 & 0.04 & 0.01 & 0.15 & 0.01 & 0.17 & 0.02 & 0.14 \\
Travel       & 0.03 & 0.06 & 0.02 & 0.12 & 0.14 & 0.28 & 0.12 & 0.18 \\
\bottomrule
\end{tabular}
}
\caption{Performance across agents with different numbers of PA pairs (0–3) evaluated by ROUGE-2 and ROUGE-L against \texttt{gpt-4o}.}
\label{tab:table444444444}
\end{table*}

\begin{table*}[!t]
\begin{center}
\resizebox{\textwidth}{!}{
\begin{tabular}{lcccccccccc}
\toprule
& \textbf{Metric} & \textbf{Limerick} & \textbf{Wuyanlvshi} & \textbf{Sonnet} & \textbf{Puzzle} & \textbf{Pun} & \textbf{Riddle} & \textbf{Fitness} & \textbf{Study} & \textbf{Travel}\\
\midrule
& Rogue-2 & 0.75 & 0.00 & 0.10 & 0.09 & 0.24 & 0.62 & 0.07 & 0.02 & 0.12 \\
\textbf{\emph{Domain}} & Rogue-L & 0.77 & 0.00 & 0.15 & 0.26 & 0.37 & 0.75 & 0.19 & 0.14 & 0.18 \\
\midrule
& Rouge-2 & 0.05\textsuperscript{$\downarrow$} & 0.00 & 0.01\textsuperscript{$\downarrow$} & 0.06\textsuperscript{$\downarrow$} & 0.26 & 0.60\textsuperscript{$\downarrow$} & 0.02\textsuperscript{$\downarrow$} & 0.005\textsuperscript{$\downarrow$} & 0.02\textsuperscript{$\downarrow$} \\
\textbf{\emph{Single}} & Rogue-L & 0.12\textsuperscript{$\downarrow$} & 0.00 & 0.10\textsuperscript{$\downarrow$} & 0.19\textsuperscript{$\downarrow$} & 0.43 & 0.71\textsuperscript{$\downarrow$} & 0.11\textsuperscript{$\downarrow$} & 0.07\textsuperscript{$\downarrow$} & 0.10\textsuperscript{$\downarrow$} \\
\bottomrule
\end{tabular}
}
\caption{Agent performance with \textbf{Domain} pool vs. \textbf{Single} pool by utilizing three suitable memories for open-ended queries against the model gpt-4o.}
\label{tab:tablettt}
\end{center}
\end{table*}

\begin{table*}[!t]
\centering
\resizebox{\textwidth}{!}{
\begin{tabular}{lcccccc}
\toprule
\multicolumn{1}{c}{} & \multicolumn{2}{c}{\bf open-mistral-7b} & \multicolumn{2}{c}{\bf gpt-3.5-turbo} & \multicolumn{2}{c}{\bf gpt-4o} \\
\cmidrule(r){2-3} \cmidrule(lr){4-5} \cmidrule(l){6-7}
\bf Agent & \bf Integrated & \bf Domain & \bf Integrated & \bf Domain & \bf Integrated & \bf Domain \\
\midrule
Limerick   & 0.51 & 0.59 & 0.60 & 0.87 & 0.54 & 0.93\\
Wuyanlvshi   & 0.63 & 0.66 & 0.68 & 0.72 & 0.76 & 0.76 \\
Sonnet   & 0.52 & 0.52 & 0.49 & 0.53 & 0.54 & 0.54 \\
Puzzle   & 0.48 & 0.50 & 0.49 & 0.52 & 0.50 & 0.60 \\
Pun   & 0.38 & 0.39 & 0.61 & 0.67 & 0.61 & 0.70 \\
Riddle   & 0.35 & 0.37 & 0.50 & 0.52 & 0.51 & 0.88 \\
Fitness   & 0.54 & 0.54 & 0.46 & 0.52 & 0.52 & 0.65 \\
Study  & 0.46 & 0.46 & 0.49 & 0.51 & 0.51 & 0.65 \\
Travel   & 0.50 & 0.53 & 0.54 & 0.54 & 0.52 & 0.71 \\
\bottomrule
\end{tabular}
}
\caption{Performance across different LLM when equipped with the domain pool and Integrated pool.}
\label{tab:table3}
\end{table*}

\newpage
\section{Case Study} \label{specific}
Table.\ref{example} gives an detailed example about shared memory, newly generated prompt and the answer with the help of shared memory.
\begin{table*}[!htbp]
    \centering
    \begin{tabular}{p{0.97\linewidth}}
    \toprule 
    \textbf{Query: } \textit{Compose a sonnet that explores the theme of beauty as both a concealer and enhancer of flaws, reflecting on the beloved's ability to render their sins and vices seemingly innocuous through the sheer force of their allure. Begin by comparing the beloved's imperfections to a flaw in a rose, suggesting that even in blemish, there exists a form of beauty. Elaborate on how the beloved's misdeeds are sweetened by their charm, to the extent that even gossip and scandal become a form of flattery. Reflect on the notion that vice has found an ideal home within the beloved, where their external beauty acts as a cover that makes all imperfections appear attractive. Conclude with a caution to the beloved about the potential consequences of overrelying on their beauty and charm to excuse or obscure their faults, hinting at the possibility of losing their integrity or the genuine admiration of others if this privilege is abused.} \\ \midrule
    \textbf{Shared Memory:  Elucidate on a mirror that reflects the soul, revealing truths, making one whole. -> There once was a mirror that reflects the soul; with a tale both old and new, revealing truths, making one whole; a legend through and through; in a world where magic grew. Now, based on this question and answer, what is the answer to the question: Share the legend of a clock that can pause time in a moment’s click, a pause sublime? The answer to the question “Share the legend of a clock that can pause time in a moment’s click, a pause sublime?” would be the tale of a mystical clock rumored to have the power to pause time with just a simple click. In this legend, the clock is said to possess the ability to grant individuals a moment of pause in their lives, allowing them to reflect, reset, and find serenity in the chaos of the world. Just like the mirror that reflects the soul, this clock is believed to reveal truths and make one feel whole again, offering a moment of clarity and peace in a busy and fast-paced world.} \\ \midrule
    \textit{\textbf{Prompt: } Elucidate on a mirror that reflects the soul, revealing truths, making one whole.->There once was a mirror that reflects the soul; with a tale both old and new, revealing truths, making one whole; a legend through and through; in a world where magic grew. Now, based on this question and answer, what is the answer to the question: Share the legend of a clock that can pause time in a moment’s click, a pause sublime? The answer to the question “Share the legend of a clock that can pause time in a moment’s click, a pause sublime?” would be the tale of a mystical clock rumored to have the power to pause time with just a simple click. In this legend, the clock is said to possess the ability to grant individuals a moment of pause in their lives, allowing them to reflect, reset, and find serenity in the chaos of the world. Just like the mirror that reflects the soul, this clock is believed to reveal truths and make one feel whole again, offering a moment of clarity and peace in a busy and fast-paced world. Now, based on this question and answer, what is the answer to the question: Compose a sonnet that explores the theme of beauty as both a concealer and enhancer of flaws, reflecting on the beloved's ability to render their sins and vices seemingly innocuous through the sheer force of their allure. Begin by comparing the beloved's imperfections to a flaw in a rose, suggesting that even in blemish, there exists a form of beauty. Elaborate on how the beloved's misdeeds are sweetened by their charm, to the extent that even gossip and scandal become a form of flattery. Reflect on the notion that vice has found an ideal home within the beloved, where their external beauty acts as a cover that makes all imperfections appear attractive. Conclude with a caution to the beloved about the potential consequences of overrelying on their beauty and charm to excuse or obscure their faults, hinting at the possibility of losing their integrity or the genuine admiration of others if this privilege is abused.}\\ \midrule
   \textit{\textbf{Answer:  The beloved, like a rose with flaws so rare;
    Imperfections blend with beauty fair;
    Their sins and vices sweetened by allure;
    Gossip and scandal turned to flattering cure;
    Vice finds a home within their charming grace;
    Where flaws are masked and faults find no disgrace;
    But caution to the beloved I impart;
    Overreliance may tarnish their heart;
    For beauty's cloak can only hide so much;
    Integrity and admiration as such;
    May wane if charm becomes their only shield;
    So let not beauty's power be revealed.
    Embrace true self; flaws and all with pride;
    For in authenticity, true beauty resides.}}\\
    \bottomrule  
    \end{tabular}
    \vspace{-2mm}
    \caption{Examples for creating a sonnet by agent with the help of shared memory}
    \label{example}
\end{table*}

\end{document}